\documentclass[lettersize,journal]{IEEEtran}
\usepackage{amsmath,amsfonts}
\usepackage{algorithmic}
\usepackage{algorithm}
\usepackage{array}
\usepackage[caption=false,font=normalsize,labelfont=sf,textfont=sf]{subfig}
\usepackage{textcomp}
\usepackage{stfloats}
\usepackage{url}
\usepackage{verbatim}
\usepackage{graphicx}
\usepackage{cite}
\usepackage{booktabs}
\usepackage{pifont}
\usepackage{xcolor}

\usepackage{arydshln}

\begin{document}

\title{GOPI: Generation-Oriented 3D Pose Inference for Furniture Insertion from Single-View RGB-D Indoor Scenes}

\author{Ruifeng Zhai, Renjie Liu, Guangrun Wang, Liang Lin$^{*}$
\thanks{$^*$Corresponding author: Liang Lin. Ruifeng Zhai and Renjie Liu are with the School of Computer Science and Engineering, Sun Yat-sen University, Guangzhou 510000, China (e-mail: zhairf@mail.sysu.edu.cn
; liurj26@mail2.sysu.edu.cn
). Guangrun Wang and Liang Lin are with the School of Computer Science and Engineering, Sun Yat-sen University, Guangzhou 510000, China; the Guangdong Key Laboratory of Big Data Analysis and Processing; and X-Era AI Lab (e-mail: wanggrun@gmail.com
; linliang@ieee.org
)
}
}

\maketitle

\begin{abstract}
We study the problem of inserting new furniture into indoor scene images. 
Under masked single-view 2D image-plane conditioning, however, the physical scale of the inserted furniture relative to the scene cannot be uniquely determined, making physically grounded furniture placement underdetermined from image evidence alone. 
We therefore reformulate the task as a combination of 3D pose inference and geometry-guided image generation, where estimating a geometrically plausible 3D placement is essential for reliable synthesis.

To this end, we propose a two-stage framework. 
For 3D placement, we introduce GOPI, a generation-oriented 3D pose inference framework that addresses the underdetermined nature of single-view furniture insertion through data-driven iterative inference, producing geometrically plausible object placements. 
For image generation, we develop a geometry-guided conditioning strategy that projects the inferred 3D pose into the image plane as a pixel-aligned constraint, enforcing consistency between the synthesized image and the underlying 3D geometry.

Experiments validate the proposed framework from both 3D placement
and image generation perspectives.
For 3D placement, GOPI achieves higher overall feasibility than
direct regression and an adapted PlaceIt3D baseline, with stronger
collision avoidance, insertion-region compliance, and reference-pose
accuracy. Its effectiveness extends to unseen scenes, replacement
furniture, and independently specified insertion regions.
For image generation, the inferred placements improve geometric
control across furniture scales while maintaining overall image quality
comparable to a matched 2D-only baseline.

\end{abstract}


\section{Introduction} \label{sec:Introduction}
\IEEEPARstart{I}{n} recent years, content insertion techniques based on image generation and editing have made remarkable progress~\cite{Rombach_2022_CVPR, podell2023sdxlimprovinglatentdiffusion}. 
In particular, mask-conditioned and instruction-guided generation methods can produce visually plausible insertions directly in the image plane~\cite{Lugmayr_2022_CVPR,couairon2022diffeditdiffusionbasedsemanticimage,Xie_2023_CVPR,10.1145/3581783.3612200}. 
However, these approaches primarily assess correctness through appearance consistency under a given viewpoint, without explicitly modeling the physical scale or three-dimensional relationship between the inserted object and the surrounding scene. 
This limitation becomes especially critical for indoor furniture insertion, where plausibility depends not only on appearance realism but also on geometrically valid placement.

\begin{figure}[t]
    \centering
    \includegraphics[width=1.0\linewidth]{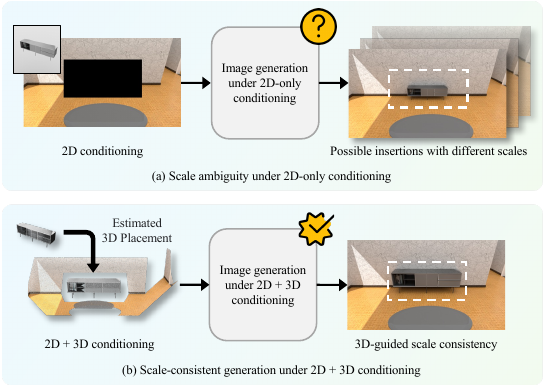}
    \caption{
    \textbf{Ambiguity under 2D vs. scale consistency with 3D guidance.}
    (a) Under 2D-only conditioning, the same masked input can correspond to multiple geometrically plausible insertion results with different object scales, due to inherent scale ambiguity.
    (b) By introducing 3D reasoning, the ambiguity is reduced, leading to scale-consistent image generation that aligns with the underlying 3D geometry.
    }
    \label{fig:scale}
\end{figure}

Furniture differs fundamentally from clothing and other appearance-dominated objects: it is rigid, strongly scale-sensitive, and tightly constrained by scene geometry. 
Given only a 2D image and an insertion mask, the observation mainly determines the image-plane extent of the inserted object, while its physical size and placement scale in 3D remain ambiguous. 
As a result, when the imaging conditions of the inserted furniture differ from those of the target environment, the inserted object may look visually reasonable yet still be physically implausible in scale relative to the scene (Fig.~\ref{fig:scale}\,(a)). 
Such ambiguity is not merely cosmetic; it can mislead users about furniture size and compromise downstream real-world decision making.

Recent progress in generative modeling further suggests that geometric coherence often requires reasoning over an explicit or implicit 3D representation. 
For example, text-to-3D optimization methods~\cite{poole2022dreamfusiontextto3dusing2d,Lin_2023_CVPR,NEURIPS2023_1a87980b} and multi-view diffusion models~\cite{Liu_2023_ICCV,shi2024mvdreammultiviewdiffusion3d,Long_2024_CVPR} leverage 3D priors to improve cross-view consistency. 
Meanwhile, 3D indoor layout and rearrangement methods~\cite{9665852,paschalidou2021atissautoregressivetransformersindoor,Tang_2024_CVPR,wei2023legonetlearningregularrearrangements} provide powerful mechanisms for object placement in structured 3D scenes. 
However, applying these advances to furniture insertion from masked
single-view RGB-D observations requires translating approximate
image-space insertion intent into a metric 3D placement under
partial observability and conveying that placement to image synthesis.

Motivated by this gap, we argue that reliable furniture insertion should not be formulated as a purely 2D image generation problem under masked single-view conditioning.
To this end, we build a pose-first furniture insertion framework that integrates established point-cloud representations, iterative refinement techniques, and image-generation architectures with task-specific geometric modeling.
The framework first infers a 3D furniture placement from partial single-view RGB-D observations and then uses this placement to guide image synthesis.

For placement, we propose \textbf{GOPI}, which represents approximate mask-specified insertion intent as a 3D frustum.
By jointly considering furniture geometry and the observed scene, GOPI refines the furniture pose through frustum-aware geometric representations and consistency constraints.
For generation, we render the estimated metric placement as a geometry-aligned projection and use it as an additional spatial condition, transferring the inferred 3D location, orientation, and scale to the appearance synthesis process.

Experiments under both asset-disjoint and scene-disjoint splits show that GOPI achieves higher overall placement feasibility than PTv3 Regression and an adapted PlaceIt3D baseline.
For image generation, projection conditioning improves projection--generation alignment over a 2D-only ablation and pretrained AnyDoor across all tested furniture scales, with complementary evidence from human evaluation.

Our contributions are three-fold:
\begin{itemize}
    \item We formulate mask-conditioned furniture insertion as
    a pose-first task under partial single-view RGB-D observations,
    representing approximate image-space insertion intent as
    a 3D frustum.

    \item We propose GOPI, a generation-oriented pose inference
    method that combines furniture and observed-scene geometry
    with frustum-aware representations and consistency constraints
    for iterative furniture placement refinement.

    \item We integrate the estimated metric placement with image
    synthesis through a geometry-aligned projection condition.
    Extensive evaluations assess placement feasibility,
    reference-pose accuracy, and projection--generation alignment,
    complemented by human evaluation.
\end{itemize}

\section{Related Work}
\label{sec:Related_Work}

\paragraph{2D Image Editing and Reference-Based Insertion.}
Diffusion-based latent generative models~\cite{Rombach_2022_CVPR,podell2023sdxlimprovinglatentdiffusion},
together with classifier-free guidance~\cite{ho2022classifierfreediffusionguidance},
support controllable image synthesis.
Mask-guided and instruction-based editing methods~\cite{Lugmayr_2022_CVPR,couairon2022diffeditdiffusionbasedsemanticimage,Xie_2023_CVPR,10.1145/3581783.3612200,Corneanu_2024_WACV,wang2023instructeditimprovingautomaticmasks,10.1145/3664647.3680830,yuan2025flexeditmarryingfreeshapemasks}
enable localized content generation, while segmentation
priors~\cite{yu2023inpaintanythingsegmentmeets},
layout grounding~\cite{Li_2023_CVPR},
structural guidance~\cite{Zhang_2023_ICCV}, and
instruction-following objectives~\cite{Brooks_2023_CVPR}
provide additional control.
Virtual try-on methods~\cite{choi2024improving,chong2025catv2tontamingdiffusiontransformers,chong2025catvton}
further demonstrate reference-conditioned appearance synthesis
in a specialized application.
Reference-based insertion methods more directly address the task
of composing a specified object into a target scene.
AnyDoor~\cite{anydoor2024} combines identity features and detail
information to insert reference objects at user-specified locations.
ObjectMate~\cite{objectmate2025} exploits recurring object instances
across images to construct paired training data and supports insertion
from one or multiple reference views.
These methods emphasize identity preservation and visual coherence,
without explicitly estimating a metric 3D placement.
GOPI instead infers furniture placement from RGB-D observations
and uses its projection to guide generation.
We evaluate the benefit of this geometric condition through
comparisons with pretrained AnyDoor and a 2D-only ablation
using the same generation backbone architecture and editing masks.

\paragraph{Semantic and Geometric Object Placement.}
Recent methods study where a new object can be placed within
an existing scene.
Seeing the Unseen~\cite{seeingunseen2024} predicts semantically
appropriate image regions for an absent object given an image
and an object category.
It addresses placement from incomplete visual evidence, but outputs
a 2D region rather than a metric object pose.
FirePlace~\cite{fireplace2025} combines multimodal language-model
reasoning with geometric constraint construction and solving
for object placement in existing 3D scenes.
PlaceIt3D~\cite{placeit3d2025} studies language-guided placement
given a reconstructed scene point cloud, a 3D asset, and an instruction,
requiring semantic, geometric, and physical reasoning.
GOPI shares their goal of plausible new-object placement,
but uses a coarse image-space mask to specify insertion intent
under partial single-view RGB-D observations.
It encodes this constraint as a 3D frustum and connects pose
inference to downstream image generation.
We include an adapted PlaceIt3D baseline to compare placement
methods under our task setting.
This task also differs from RGB-D pose estimation and CAD
alignment~\cite{Avetisyan_2019_CVPR,xiang2018posecnnconvolutionalneuralnetwork,Wang_2019_CVPR,Qi_2019_ICCV,Hou_2019_CVPR},
which localize or align objects already present in the observation.

\paragraph{3D Layout and Geometry-Consistent Generation.}
Indoor layout methods, including 
SceneFormer~\cite{9665852},
ATISS~\cite{paschalidou2021atissautoregressivetransformersindoor},
DiffuScene~\cite{Tang_2024_CVPR},
Mixed Diffusion~\cite{hu2024mixeddiffusion3dindoor}, and
SemLayoutDiff~\cite{sun2025semlayoutdiffsemanticlayoutgeneration},
model structured configurations of scene objects.
LEGO-Net~\cite{wei2023legonetlearningregularrearrangements}
introduces iterative rearrangement under learned structural priors;
we adapt its refinement schedules to our insertion setting.
Text-to-3D methods~\cite{poole2022dreamfusiontextto3dusing2d,Lin_2023_CVPR,NEURIPS2023_1a87980b}
and multi-view generators~\cite{Liu_2023_ICCV,liu2024syncdreamergeneratingmultiviewconsistentimages,shi2024mvdreammultiviewdiffusion3d,Long_2024_CVPR}
further explore geometric consistency through 3D representations
and cross-view modeling.
These works provide relevant foundations, while GOPI focuses on
placing a given furniture asset under partial observations
and conveying its inferred geometry to image synthesis.
Our synthesis stage builds on IDM-VTON~\cite{choi2024improving},
adapted with a geometry-aligned projection as an additional
spatial condition.

\begin{figure*}[t]
    \centering
    \includegraphics[width=1.0\linewidth]{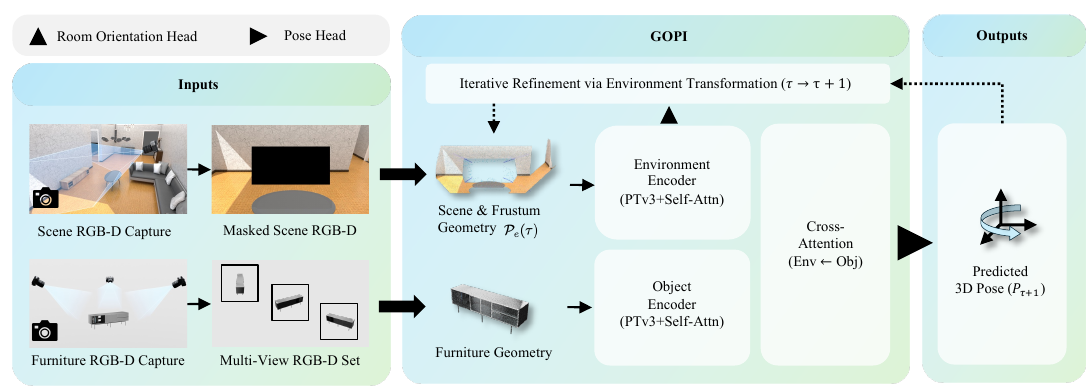}
    \caption{
    \textbf{Overview of the furniture insertion task and the GOPI framework.}
    The inputs consist of a masked single-view scene RGB-D observation, obtained by capturing an indoor scene and specifying a user mask, together with a multi-view RGB-D set of the target furniture.
    Based on these inputs, GOPI constructs scene--frustum geometry and furniture geometry, encodes them with environment and object encoders, performs cross-stream interaction for pose prediction, and iteratively updates the environment representation for refinement.
    The output is a plausible 3D furniture pose, including position and yaw, for downstream image generation.
    }
    \label{fig:pipeline}
\end{figure*}

\section{Problem Formulation} \label{sec:problem_formulation}
In this section, we formalize furniture placement under a single-view, mask-conditioned RGB-D setting, specifying the task objective, input-output definition, and the intrinsic ambiguity induced by partial observability (Fig.~\ref{fig:pipeline}).

\subsection{Task Definition}
\label{sec:task_definition}
We study furniture placement in indoor scenes under a generation-oriented setting, where the goal is to infer a plausible 3D pose for inserting a new furniture item into an existing environment. 
The predicted pose is expected to support downstream image generation while remaining geometrically feasible and semantically plausible in the scene context.
Let $\mathcal{S}$ denote an indoor scene and $\mathcal{O}$ denote a target furniture object with fixed three-dimensional geometry. 
The task is to infer a 3D pose of $\mathcal{O}$ within $\mathcal{S}$.

A desirable furniture pose should satisfy three high-level considerations. 
First, it should be \emph{geometrically feasible}, satisfying basic constraints such as non-penetration with the environment and proper support from the ground. 
Second, it should be \emph{human-plausible}, aligning with common furniture placement patterns and usage preferences observed in real indoor environments. 
Third, it should be \emph{intent-consistent}, i.e., compatible with a user-specified target region or intended placement context within the scene.

\subsection{Inputs and Outputs}
\label{sec:inputs_outputs}
We consider furniture placement from partial single-view observations and user-provided insertion constraints, with the goal of inferring a generation-oriented pose for downstream image synthesis.

\paragraph{Environment Observation and Insertion Constraint.}
The indoor scene $\mathcal{S}$ is observed under a single-view RGB-D acquisition with known camera parameters. 
Given a user-specified two-dimensional insertion region $M$ on the image plane, the RGB image and depth map are partially masked within $M$, with values inside $M$ removed and treated as invalid. 
We denote the masked scene RGB-D observation as
$
\mathcal{I}_s = \{I_s, D_s, K_s, T_s\},
$
where $I_s$ and $D_s$ denote the masked RGB image and depth map, respectively, with entries inside $M$ treated as invalid, while $K_s$ and $T_s$ denote the camera intrinsics and extrinsics, respectively.

The mask $M$ indicates a coarse target area for placing the new furniture. 
It is not required to tightly match the true image projection of the inserted object, and may include surrounding context, reflecting the fact that user intent is approximate rather than geometrically precise.

\paragraph{Furniture Geometry Prior.}
The target object $\mathcal{O}$ is associated with a geometric prior obtained from multi-view furniture RGB-D observations, denoted as
$\mathcal{I}_o = \{ I_o^{(v)}, D_o^{(v)}, K_o^{(v)}, T_o^{(v)} \}_{v=1}^V$. 
Here, $T_o^{(v)}$ maps the object-centric coordinate frame to the camera frame of view $v$.
These observations are used to reconstruct the three-dimensional geometry of $\mathcal{O}$ and to define its local object-centric coordinate system. 
The furniture geometry is assumed to be rigid and fixed during pose inference.

\paragraph{Output Pose.}
The output of the system is a three-dimensional pose of the target furniture $\mathcal{O}$ expressed in the coordinate frame of the scene $\mathcal{S}$. 
Under the upright scene assumption, we represent the furniture pose as
$\mathbf{p} = (\mathbf{t}, \theta)$,
where $\mathbf{t} \in \mathbb{R}^3$ denotes the translation and $\theta$ denotes a periodic yaw angle with respect to the gravity-aligned vertical axis.

\paragraph{Assumptions.}
Unless otherwise specified, we assume an upright scene configuration in which the gravity direction is known and aligned with the vertical axis.

\subsection{Observability and Ambiguity}
\label{Observability&Ambiguity}
Furniture placement in indoor scenes is constrained by functional role, spatial relationships with surrounding objects, and the global room layout. 
Under full observability, these cues can substantially narrow the set of plausible placements.

In our setting, however, the scene is observed from only a single view, and the user-specified insertion mask removes the local geometry in the target region. 
As a result, contextual cues such as nearby objects, wall proximity, and detailed layout structure are only partially observed or entirely missing, which significantly enlarges the ambiguity of the solution space.

Importantly, this ambiguity is intrinsic to the observation setting rather than an artifact of annotation or model design. 
Multiple furniture poses may remain compatible with the same observed evidence and user intent, while leading to different yet plausible spatial configurations. 
Therefore, rather than assuming a uniquely determined pose, our goal is to infer a geometrically reasonable placement that is suitable for downstream image generation.

\section{Method} \label{sec:Method}
We propose GOPI, a generation-oriented framework that infers a geometrically plausible 3D furniture pose from a masked single-view RGB-D scene and multi-view RGB-D observations of the target furniture (Fig.~\ref{fig:pipeline}). 
The method consists of three components: geometric representations of the observed scene, insertion region, and target object; a dual-stream network for object--environment interaction and pose prediction; and an iterative refinement scheme that progressively improves geometric consistency under partial observability.

\subsection{Geometric Representations}
\label{sec:scene_representation}

\paragraph{Masked Scene Geometry Representation.}
From the masked scene RGB-D observation $\mathcal{I}_s$, we back-project valid depth pixels outside the insertion region $M$ using the camera parameters $K_s$ and $T_s$ to obtain a partial scene point cloud $\mathcal{P}_s = \{ \mathbf{x}_i \in \mathbb{R}^3 \}$. 
This representation captures the observed scene geometry while explicitly excluding the unknown geometry inside the masked insertion region.

\paragraph{Insertion Frustum Representation.}
The image-plane insertion mask $M$ induces a frustum-based geometric reference in scene space.
We construct a truncated camera frustum by back-projecting the rectangular bounding box of $M$ through the camera intrinsics $K_s$, with a fixed near depth $z_n$ and far depth $z_f=z_n+L$.
Instead of treating the frustum as a closed volume, we sample 3D points on its lateral surfaces and edges, obtaining a frustum boundary point set $\mathcal{P}_f=\{\mathbf{x}_i\in\mathbb{R}^3\}$.
This open geometric scaffold encodes where insertion is intended to occur, while remaining agnostic to the unobserved geometry inside the masked region.

\paragraph{Furniture Geometry Representation.}
We reconstruct the furniture geometry as a point cloud $\mathcal{P}_o = \{ \mathbf{x}_i \in \mathbb{R}^3 \}$ by back-projecting and aggregating depth pixels from the multi-view RGB-D observations $\mathcal{I}_o$ into a common object-centric coordinate frame.
Following the 3D-FRONT convention, the local origin is defined at the center of the bottom contact region, with the vertical axis aligned to gravity.
This representation serves as the geometric prior of the target furniture throughout pose inference.

\subsection{GOPI Network Architecture}
\label{sec:network_architecture}
\paragraph{Dual-Stream Geometric Encoding.}

To reason about furniture placement, the network must jointly model the geometry of the target object and that of the surrounding environment, while preserving their distinct structural roles.
We therefore adopt a dual-stream point-based encoding architecture.
Specifically, we consider two point sets:
(i) the furniture point cloud $\mathcal{P}_o$, and
(ii) the environment point cloud $\mathcal{P}_e = \mathcal{P}_s \cup \mathcal{P}_f$,
which combines the observed scene geometry and the insertion frustum.

Each point is associated with a feature vector encoding geometric, semantic, and contextual information, including 3D coordinates, RGB color, DINOv2 embeddings, and a one-hot indicator of point type (scene, frustum, or furniture). 
For frustum boundary points, image-dependent features are unavailable and thus set to zero.
In addition, we incorporate a frustum-relative geometric descriptor defined as the signed distances to the four lateral frustum planes $\{(\mathbf{n}_j, \mathbf{q}_j)\}_{j=1}^{4}$, where $\mathbf{n}_j$ denotes the outward unit normal and $\mathbf{q}_j$ is a point on each plane:
\begin{equation}
d_j(\mathbf{x}) = \langle \mathbf{n}_j,\; \mathbf{x} - \mathbf{q}_j \rangle.
\label{eq:frustum_signed_distance}
\end{equation}
This descriptor provides a compact encoding of point--frustum spatial relationships, enabling geometry-aware reasoning under partial observability.

All per-point attributes are linearly projected and concatenated into unified point features.
The furniture point set $\mathcal{P}_o$ and the environment point set $\mathcal{P}_e$ are processed by two parallel PTv3~\cite{PTv3} encoders with the same architecture but independent parameters, enabling object-centric and environment-centric representations to be learned separately.
The resulting features are refined by self-attention within each stream, yielding $\mathbf{F}_o \in \mathbb{R}^{N_o \times C}$ and $\mathbf{F}_e \in \mathbb{R}^{N_e \times C}$, where $N_o$ and $N_e$ denote the numbers of downsampled furniture and environment points, and $C$ is the feature dimension.

\paragraph{Cross-Stream Interaction.}
To enable environment-aware placement reasoning, we apply cross-attention where the environment features $\mathbf{F}_e$ query the furniture features $\mathbf{F}_o$ as keys and values.
The resulting cross-attended environment features $\tilde{\mathbf{F}}_e$ encode furniture-conditioned spatial cues for subsequent pose prediction.
This interaction allows each environment location to be interpreted in the context of the target furniture geometry, which is crucial for placement reasoning under partial observability.

\paragraph{Pose Readout.}
We obtain a compact representation for pose prediction by applying generalized mean (GeM) pooling over the cross-attended environment features.
Specifically, we aggregate $\tilde{\mathbf{F}}_e$ into a feature vector.

The pooled feature is then passed through two lightweight MLP heads with identical architectures but independent parameters to predict a forward furniture pose update
$\hat{\delta}=(\mathbf{t},\mathbf{u}_\theta)$,
where $\mathbf{t}\in\mathbb{R}^3$ denotes the predicted translation increment of the furniture and
$\mathbf{u}_\theta=(\cos\theta,\sin\theta)\in\mathbb{R}^2$
parameterizes its predicted yaw increment.

\paragraph{Room Orientation Readout.}
Indoor furniture orientations are known to be strongly correlated with the dominant room axes, as most objects in indoor layouts follow a Manhattan-style structure.
This phenomenon has been empirically observed in large-scale indoor datasets, e.g., SemLayoutDiff~\cite{sun2025semlayoutdiffsemanticlayoutgeneration} reports that the vast majority of furniture orientations in 3D-FRONT are aligned with the coordinate axes.
Motivated by this structural regularity, we additionally estimate a room orientation as a global structural cue for pose inference.

Specifically, we apply GeM pooling over $\mathbf{F}_e$ to obtain a compact global representation, which is passed to a lightweight MLP head to predict a room orientation vector
$\hat{\mathbf{u}}_{\phi} = (\cos\phi, \sin\phi)$,
where $\phi$ denotes the yaw angle of a dominant room axis.

\paragraph{Geometric Prediction Formulation.}
We summarize the network as a learned geometric mapping from the object–environment configuration to pose and room orientation predictions.
Given the furniture point cloud $\mathcal{P}_o$ and the environment point cloud
$\mathcal{P}_e$, together with their associated per-point features, the network implements a function
\begin{equation}
\mathcal{F}_{\Theta} : (\mathcal{P}_o, \mathcal{P}_e)
\;\longrightarrow\;
\big(\hat{\delta}, \hat{\mathbf{u}}_{\phi}\big),
\end{equation}
where $\Theta$ denotes the learnable parameters.

\subsection{Iterative Refinement via Environment Transformation.}
\label{sec:iterative_inference}

\paragraph{Motivation under Partial Observability.}
As discussed in Section~\ref{Observability&Ambiguity}, furniture pose inference from masked single-view RGB-D observations is intrinsically ambiguous.
Under such limited observability, one-shot pose regression is often brittle and may lead to geometrically implausible placements.
We therefore formulate pose inference as an iterative refinement process that progressively improves geometric consistency and physical feasibility.

\paragraph{Refinement Formulation.}
We formulate pose inference as iterative refinement in a furniture-centric coordinate frame.
Instead of updating the furniture pose directly, we keep the furniture geometry fixed in its canonical frame and apply the inverse of the predicted pose increment to the environment point cloud.
Starting from an initialization, the environment is iteratively transformed to refine the relative configuration between the furniture and the scene.

This formulation keeps the furniture geometry in a stable canonical frame, reduces ambiguity caused by object symmetries, and makes object--scene geometric relationships easier to measure in the transformed environment coordinates.

\paragraph{Noise-Perturbed Training for Pose Refinement.}
Given a reference furniture--environment configuration, we express the relative object--scene geometry in a canonical furniture-centric coordinate frame.
Specifically, both the furniture and the environment are transformed into the local coordinate system of the target furniture, such that the furniture point cloud $\mathcal{P}_o$ is centered at the origin.
The corresponding transformed environment point cloud is denoted by $\mathcal{P}_e^{\mathrm{ref}}$.

To train the network to predict forward furniture pose updates, we generate perturbed training states by applying random rigid transformations to the environment while keeping the furniture fixed.
We sample a perturbation $\delta = (\Delta \mathbf{t}, \Delta \theta)$, where $\Delta \mathbf{t} \in \mathbb{R}^3$ and $\Delta\theta$ is a scalar yaw perturbation, and construct
\begin{equation}
\mathcal{P}_e(\delta)
=
\left\{
\mathbf{R}(\Delta\theta)\,\mathbf{y} + \Delta\mathbf{t}
\;\middle|\;
\mathbf{y} \in \mathcal{P}_e^{\mathrm{ref}}
\right\},
\end{equation}
where $\mathbf{R}(\Delta\theta)$ denotes rotation about the gravity-aligned vertical axis.
The perturbation is sampled from Gaussian distributions over translation and yaw.
For yaw regression, the target is represented as $\mathbf{u}_{\Delta\theta}=(\cos\Delta\theta,\sin\Delta\theta)$.

Given the perturbed configuration
$(\mathcal{P}_o,\mathcal{P}_e(\delta))$,
the network predicts a forward furniture pose update
$\hat{\delta}$, whose inverse is applied to the environment
during refinement.
The training objective is defined as
\begin{equation}
\mathcal{L}_{\mathrm{ref}}
=
\mathbb{E}_{\delta}
\left[
\mathcal{L}_{t}
+
\mathcal{L}_{\theta}
+
\lambda_{\mathrm{fru}}\mathcal{L}_{\mathrm{fru}}
+
\lambda_z\mathcal{L}_z
+
\lambda_{\mathrm{room}}\mathcal{L}_{\mathrm{room}}
\right],
\end{equation}
where
$\mathcal{L}_{t}=\ell_t(\mathbf{t},\Delta\mathbf{t})$
is the translation regression loss,
$\mathcal{L}_{\theta}$ is the cosine loss between
$\mathbf{u}_\theta$ and $\mathbf{u}_{\Delta\theta}$,
$\mathcal{L}_{\mathrm{fru}}$ enforces frustum consistency,
$\mathcal{L}_z$ penalizes support misalignment, and
$\mathcal{L}_{\mathrm{room}}$ regularizes the predicted dominant room axis.

\paragraph{Frustum Consistency Constraint.}
To enforce compatibility with the image-plane insertion region, we introduce a frustum consistency loss on the transformed furniture geometry.
Given a predicted pose update $\hat{\delta}$, each furniture point $\mathbf{x}_i \in \mathcal{P}_o$ is transformed as
$\mathbf{x}_i(\hat{\delta}) = \mathbf{R}(\theta)\mathbf{x}_i + \mathbf{t}$.
Using the frustum-relative signed distances in Eq.~\ref{eq:frustum_signed_distance}, we define the point-wise frustum violation as
\begin{equation}
\tilde{v}_i(\hat{\delta})
=
\max\!\Bigl(0,\;
\max_{j \in \{1,\dots,4\}} d_j\!\left(\mathbf{x}_i(\hat{\delta})\right)
- m
\Bigr),
\end{equation}
where $m \ge 0$ is a hinge margin.

Points outside the frustum are divided into two groups:
(i) visible overflow points $\mathcal{V}_{\mathrm{vis}}$, which are not occluded by the scene, and
(ii) invisible overflow points $\mathcal{V}_{\mathrm{invis}}$, which are either occluded or projected outside the image plane.
This distinction reflects the fact that visible violations are directly observable, whereas invisible violations are less reliable as supervision signals.

Visible overflow points are penalized with a fixed weight $w_{\mathrm{vis}}$, while invisible overflow points are modulated by an adaptive weight $w_{\mathrm{invis}}$ based on the projected coverage ratio.
This design assigns stronger penalties to directly observable frustum violations and reduces the influence of invisible violations when the predicted projection sufficiently covers the reference insertion region.
The detailed definition of $w_{\mathrm{invis}}$ is provided in the supplementary material.

The final frustum consistency loss is
\begin{equation}
\mathcal{L}_{\mathrm{fru}}=
w_{\mathrm{vis}}
\frac{1}{|\mathcal{V}_{\mathrm{vis}}|}
\sum_{i \in \mathcal{V}_{\mathrm{vis}}} \tilde{v}_i(\hat{\delta}) 
+
w_{\mathrm{invis}}
\frac{1}{|\mathcal{V}_{\mathrm{invis}}|}
\sum_{i \in \mathcal{V}_{\mathrm{invis}}} \tilde{v}_i(\hat{\delta}).
\end{equation}
If either set is empty, the corresponding term is set to zero.

\paragraph{Support Alignment Loss.}
We introduce a support alignment loss that places additional emphasis on the vertical component of translation regression:
\begin{equation}
\mathcal{L}_{z}
=
\ell_t
\left(
t_z,
\Delta t_z
\right),
\end{equation}
where ${t}_z$ and $\Delta{t}_z$ denote the vertical components of the predicted and sampled translations, respectively, and $\ell_t$ is the same regression loss used in $\mathcal{L}_t$.
This loss emphasizes vertical placement accuracy and encourages the refined placement to preserve the reference support height.

\paragraph{Room Orientation Loss.}
Indoor environments typically exhibit a Manhattan-world structure, in which dominant room orientations form a pair of orthogonal horizontal axes.
Predicting a unique axis from this set is inherently ambiguous, since multiple dominant orientations may be equally valid under the same observation.

We therefore supervise the network to predict an arbitrary dominant room axis rather than a fixed canonical one.
Let $\hat{\mathbf{u}}_{\phi} = (\cos\phi,\sin\phi)$ denote the predicted room-axis direction, and let
$\mathbf{u}_x = (\cos\phi_x,\sin\phi_x)$ and
$\mathbf{u}_y = (\cos\phi_y,\sin\phi_y)$
denote the two orthogonal ground-truth room axes on the horizontal plane.

The loss is designed to be invariant to both axis swapping and sign flipping.
We define the mismatch between the prediction and each ground-truth axis as
\begin{equation}
d_x = 1 - (\hat{\mathbf{u}}_{\phi}^{\top}\mathbf{u}_x)^2,
\qquad
d_y = 1 - (\hat{\mathbf{u}}_{\phi}^{\top}\mathbf{u}_y)^2.
\end{equation}
The final loss is a soft minimum over the two candidates:
\begin{equation}
\mathcal{L}_{\mathrm{room}}
=
-\tau \log
\left(
\exp\!\left(-\frac{d_x}{\tau}\right)
+
\exp\!\left(-\frac{d_y}{\tau}\right)
\right),
\end{equation}
where $\tau > 0$ controls the sharpness of the soft minimum.

\paragraph{Iterative Inference Procedure.}
At inference time, we initialize the refinement with a reference pose
$\mathbf{p}_0=(\mathbf{c}_f,\theta_0)$,
where $\theta_0\sim\mathcal{U}(-\pi,\pi)$ and
$\mathbf{c}_f$ is the geometric center of the frustum boundary point set
$\mathcal{P}_f$.
Rather than explicitly placing the furniture at this pose, we transform the environment point cloud $\mathcal{P}_e$ into the corresponding furniture-centric frame, yielding $\mathcal{P}_e^{(0)}$.

At iteration $\tau$, we apply the learned geometric mapping
\begin{equation}
(\hat{\delta}_{\tau}, \hat{\mathbf{u}}_{\phi,\tau})
=
\mathcal{F}_{\Theta}(\mathcal{P}_o,\mathcal{P}_e^{(\tau)}),
\end{equation}
where $\hat{\delta}_{\tau}=(\mathbf{t}_\tau,\mathbf{u}_{\theta,\tau})$ is an incremental pose update predicted under the current configuration, and $\hat{\mathbf{u}}_{\phi,\tau}$ denotes the predicted dominant room axis.

Following the refinement style of LEGO-Net, we first decode the predicted yaw increment as
\begin{equation}
\theta_\tau
=
\operatorname{atan2}
\left(
[\mathbf{u}_{\theta,\tau}]_2,
[\mathbf{u}_{\theta,\tau}]_1
\right).
\end{equation}
We then construct the actual update in translation--yaw space as
\begin{equation}
\Delta_\tau
=
\alpha(\tau)
\bigl(\mathbf{t}_\tau,\theta_\tau\bigr)
+
\beta(\tau)\boldsymbol{\epsilon}_\tau,
\end{equation}
where
$\boldsymbol{\epsilon}_\tau
=
(\boldsymbol{\epsilon}^{t}_\tau,\epsilon^\theta_\tau)$
denotes zero-mean Gaussian noise in translation and yaw.
The schedules $\alpha(\tau)$ and $\beta(\tau)$ are instantiated separately for the translation and yaw components and progressively decrease during refinement.
Once the predicted translation and yaw magnitudes both fall below predefined convergence thresholds for consecutive iterations, the noise is suppressed and the refinement terminates.

We then update the environment by applying the inverse rigid transformation:
\begin{equation}
\mathcal{P}_e^{(\tau+1)}
=
\mathbf{R}(-\Delta\theta_{\tau})
\left(
\mathcal{P}_e^{(\tau)} - \Delta \mathbf{t}_{\tau}
\right),
\label{eq:iter_env}
\end{equation}
where $\Delta \mathbf{t}_{\tau}$ and $\Delta \theta_{\tau}$ denote the translational and yaw components of $\Delta_{\tau}$, respectively.

In parallel, the same increments are accumulated to obtain the final pose estimate:
\begin{equation}
\mathbf{p}_{\tau+1}
=
\mathbf{p}_{\tau}
\;\oplus\;
\Delta_{\tau},
\label{eq:iter_pose_acc}
\end{equation}
where $\oplus$ denotes pose composition.

\paragraph{Room-Axis Alignment in Refinement.}
To further stabilize yaw refinement with global structural cues, we introduce a room-axis alignment step at each iteration during the last $20\%$ of refinement iterations.
Specifically, the yaw direction $\mathbf{u}_{\theta,\tau}$ extracted from $\hat{\delta}_{\tau}$ is projected onto the nearest dominant axis induced by the predicted room orientation $\hat{\mathbf{u}}_{\phi,\tau}$, yielding an axis-aligned candidate $\tilde{\mathbf{u}}_{\theta,\tau}$.

Instead of enforcing hard snapping, we progressively blend the raw yaw prediction with its axis-aligned counterpart:
\begin{equation}
\mathbf{u}_{\theta,\tau}^{\text{final}}
=
\mathrm{normalize}\!\left(
(1-w_\tau)\,\mathbf{u}_{\theta,\tau}
+
w_\tau\,\tilde{\mathbf{u}}_{\theta,\tau}
\right),
\end{equation}

where $w_\tau$ increases linearly from $0.4$ to $0.9$
during this stage.
During this stage, we decode the blended direction as
\begin{equation}
\Delta\theta_\tau
=
\operatorname{atan2}\!\left(
[\mathbf{u}_{\theta,\tau}^{\mathrm{final}}]_2,
[\mathbf{u}_{\theta,\tau}^{\mathrm{final}}]_1
\right),
\end{equation}
and use it directly as the furniture yaw increment, without
additional step-size scaling or angular noise.
The translation increment follows the update rule defined above.

\subsection{Geometry-Guided Image Generation}
\label{sec:geo_guided_gen}

\paragraph{Geometry-aligned RGB Projection.}
Given the final refined pose $\mathbf{p}^*$, we render a geometry-aligned RGB projection of the furniture as a pixel-space conditioning signal for image generation.
Specifically, we transform the furniture point cloud $\mathcal{P}_o$ by $\mathbf{p}^*$ and render it in Blender under the same camera intrinsics and extrinsics as the observed scene:
\begin{equation}
I_o^{\mathrm{proj}} = \Pi_{\mathrm{img}}(\mathcal{P}_o \mid \mathbf{p}^*; K_s, T_s),
\end{equation}
where $\Pi_{\mathrm{img}}(\cdot)$ denotes the Blender rendering operator.
Each colored point is rendered as a small sphere, producing a pixel-aligned projection that preserves the inferred 3D location, scale, and orientation.

\paragraph{Mask Refinement under Projection.}
Since the user-specified mask provides only a coarse image-plane constraint, the projected furniture may extend beyond the original mask.
We therefore refine the generation mask by taking the axis-aligned bounding rectangle of the union between the original mask and the visible furniture projection, ensuring complete coverage of the inserted object.

\paragraph{Diffusion Conditioning with Geometric Projection.}
\begin{figure}[t]
\centering
\includegraphics[width=1.0\linewidth]{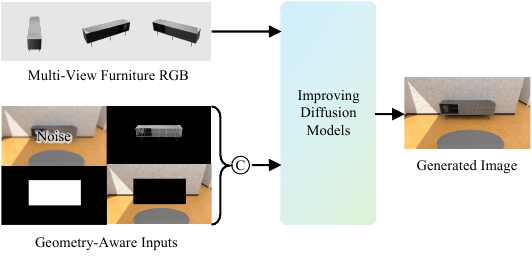}
\caption{
\textbf{Diffusion conditioning with geometry-aligned inputs.}
The geometry-aware inputs are concatenated along the channel dimension and fed into the diffusion model, where the geometry-aligned projection provides a pixel-aligned spatial constraint that anchors the generated content to the inferred 3D pose.
}

\label{fig:IDM}
\end{figure}

We adopt IDM~\cite{choi2024improving} as the diffusion backbone and incorporate geometry-consistent conditions derived from the inferred pose $\mathbf{p}^*$ (Fig.~\ref{fig:IDM}).
The geometry-aligned projection $I_o^{\mathrm{proj}}$ and the masked scene image $I_s$ are encoded into spatially aligned latent features $z_o^{\mathrm{proj}}$ and $z_s^{\mathrm{masked}}$.

The TryonNet branch takes the noisy latent $z$ together with the generation mask $m$, the masked-scene latent $z_s^{\mathrm{masked}}$, and the projection latent $z_o^{\mathrm{proj}}$.
In parallel, the GarmentNet branch encodes object-level appearance priors from the multi-view furniture observations $\{ I_o^{(v)} \}_{v=1}^V$, aggregated into a single conditioning image.
Compared with purely mask-based conditioning, $I_o^{\mathrm{proj}}$ provides a pixel-aligned spatial anchor that guides the diffusion process to preserve the inferred 3D location, scale, and orientation.

\section{Experiments}
\label{sec:Experiments}
We evaluate the proposed framework from three complementary perspectives: 
(1) geometric feasibility of the predicted 3D placements, 
(2) pose consistency with reference layouts under tolerance-based criteria, and 
(3) geometry consistency in image generation. 
Together, these evaluations assess whether the proposed pose-first formulation improves 3D placement quality and supports more geometry-consistent downstream image generation.

\subsection{Experimental Setup}

\subsubsection{Dataset}
We construct a synthetic benchmark, termed \textbf{Front3D-Insertion}, based on 3D-FRONT~\cite{fu20213dfront3dfurnishedrooms} indoor scenes and 3D-FUTURE~\cite{fu20203dfuture3dfurnitureshape} furniture assets.
For each 3D-FRONT scene, we use BlenderProc~\cite{Denninger2023} to render RGB-D observations and record the corresponding camera intrinsics and extrinsics.
To obtain diverse scene observations, camera poses are randomly sampled inside rooms subject to visibility and obstacle constraints, and each scene is rendered from up to three valid viewpoints.
Each rendered scene view has a resolution of $1920\times1080$.

To construct insertion samples, we randomly select one furniture instance from each rendered scene as the target object and use its occupied image region to define the insertion mask.
The instance mask is then converted into an axis-aligned rectangular region by enlarging its bounding rectangle with slight random padding, which better simulates user-specified insertion regions.

For each target furniture instance, we additionally render multi-view RGB-D observations in isolation using the corresponding 3D-FUTURE asset.
Each object is rendered from three canonical viewpoints at a resolution of $1024\times1024$.

Following ATISS~\cite{paschalidou2021atissautoregressivetransformersindoor}, we remove scenes with severe object interpenetration and further filter out visually implausible samples, such as those without valid target furniture or with excessively large object coverage in the image.
The final dataset contains 18,203 samples.
Our default asset-disjoint split comprises 16,676 training samples
and 1,527 test samples, with no overlap in target furniture asset IDs.
We additionally construct a scene-disjoint split by retaining the
same test set and removing all training samples from test scenes,
leaving 14,060 training samples.
This split has no overlap in either scene IDs or target asset IDs.
Unless otherwise stated, all experiments use the asset-disjoint split.

\subsubsection{Evaluation Metrics}

\paragraph{Geometric Feasibility.}
Geometric feasibility is evaluated using four criteria: support validity, collision validity, frustum validity, and overall feasibility.

\textbf{Support.}
For floor-supported objects, we measure the absolute distance between
the lowest point of the placed object and the floor, with thresholds
of 5/10/15\,cm for strict/medium/loose evaluation.
For ceiling-hung objects, we measure the absolute distance between
the highest point and the ceiling, allowing 10/15/20\% of the object
height to accommodate adjustable hanging heights.

\textbf{Collision.}
We test the placed furniture mesh for intersection with the
single-view observed-surface mesh outside the insertion mask
and room-boundary walls.
The walls are expanded outward by 5/10/15\,cm for
strict/medium/loose evaluation.
A prediction is collision-valid if neither intersection occurs.
Details are provided in the supplementary material Collision Evaluation.

\textbf{Frustum Validity.}
To measure whether the predicted placement remains inside the valid insertion region, we compare the predicted rectangle mask $\mathcal{R}_{\mathrm{pred}}$ with the reference insertion mask $\mathcal{M}_{\mathrm{ref}}$.
$\mathcal{R}_{\mathrm{pred}}$ is obtained by projecting the predicted object onto the image plane, computing the visible mask via z-buffering with the masked scene depth map, and taking its axis-aligned bounding rectangle.
The frustum violation ratio is defined as $r_{\mathrm{frus}} = |\mathcal{R}_{\mathrm{pred}} \setminus \mathcal{M}_{\mathrm{ref}}| / |\mathcal{R}_{\mathrm{pred}}|$, and a prediction is frustum-valid if $r_{\mathrm{frus}} \le \tau_{\mathrm{frus}}$.
This metric is applied only when the target furniture is kept at its dataset-provided original 3D scale. 
We set $\tau_{\mathrm{frus}}$ to $0.05$, $0.10$, and $0.15$ for strict, medium, and loose evaluation, respectively.

\textbf{Overall Feasibility.}
A prediction is feasible only if support, collision, and frustum validity are all satisfied, i.e., $\mathrm{Overall}=\mathrm{Supp}\land\mathrm{Coll}\land\mathrm{Frus}$.
Unless otherwise specified, all results are reported under the \emph{medium} threshold setting.

\paragraph{Pose Accuracy.}
We report \textbf{TransAcc@0.3m} and \textbf{YawAcc@10$^\circ$}, i.e., the percentages of predictions whose translation and yaw errors are below $0.3$ m and $10^\circ$, respectively.
Because furniture placement is inherently multi-modal, the dataset pose is treated as a reference layout rather than the only valid solution.
These tolerance-based metrics therefore measure how well the predicted pose remains consistent with human-designed reference arrangements.

\paragraph{Geometry Consistency.}
To quantify consistency between the synthesized furniture and the inferred 3D geometry in terms of scale and shape, we introduce a projection-based IoU metric, termed \textbf{Proj-Gen IoU}.

Specifically, we obtain the generated furniture mask $\mathcal{M}_{\mathrm{gen}}$ using Grounded-SAM~\cite{ren2024groundedsamassemblingopenworld} within the masked region of the synthesized image.
As the geometric reference, we extract the projection mask $\mathcal{M}_{\mathrm{proj}}$ from the rendered projection $I_o^{\mathrm{proj}}$ by thresholding valid pixels.
The metric is defined as
\begin{equation}
\mathrm{IoU}_{\mathrm{Proj-Gen}}
=
\frac{
|\mathcal{M}_{\mathrm{gen}} \cap \mathcal{M}_{\mathrm{proj}}|
}{
|\mathcal{M}_{\mathrm{gen}} \cup \mathcal{M}_{\mathrm{proj}}|
}.
\end{equation}

Although this metric can be affected by segmentation quality and rendering density, it provides a practical proxy for geometry consistency in image space.

\subsubsection{Implementation Details}
The scene point cloud is back-projected from the rendered RGB-D observation, and the target furniture point cloud is fused from multi-view RGB-D renderings. 
Both point clouds are downsampled using a $0.01$ m grid.

During training, relative pose perturbations are sampled around the target configuration, with 
$\Delta \mathbf{t} \sim \mathcal{N}(\mathbf{0}, 0.5^2 \mathbf{I})$ 
for translation and 
$\Delta \theta \sim \mathcal{N}(0, (60^\circ)^2)$ 
for yaw. 
The 3D pose inference network is trained with AdamW using a batch size of $4$ and gradient accumulation over $2$ steps.

At inference time, GOPI runs up to $50$ refinement iterations using the deterministic and stochastic update schedules described in Section~\ref{sec:iterative_inference}. 
Additional training and inference details, including schedule parameters, early stopping, room-axis alignment, and IDM training settings, are provided in the supplementary material.

\subsection{Quantitative Results on 3D Pose Estimation}

\begin{table*}[t]
\centering
\caption{
\textbf{3D placement results under asset-disjoint and scene-disjoint splits.}
$\dagger$ denotes adapted PlaceIt3D.
GT is a reference rather than a strict upper bound.
Best learned-method results within each split are bolded.
}
\label{tab:main_results}
\setlength{\tabcolsep}{6pt}
\setlength{\dashlinedash}{2pt}
\setlength{\dashlinegap}{2pt}
\renewcommand{\arraystretch}{1.1}
\begin{tabular}{@{}lcccccc@{}}
\toprule
& \multicolumn{4}{c}{\textbf{Feasibility (\%) $\uparrow$}}
& \multicolumn{2}{c}{\textbf{Accuracy (\%) $\uparrow$}} \\
\cmidrule(lr){2-5}\cmidrule(lr){6-7}
\textbf{Method}
& \shortstack{Support@M}
& \shortstack{Collision@M}
& \shortstack{Frustum@M}
& \shortstack{Overall@M}
& \shortstack{TransAcc@0.3m}
& \shortstack{YawAcc@10$^\circ$} \\
\midrule
\multicolumn{7}{c}{\textit{(a) Asset-disjoint split}} \\
\midrule[0.4pt]
GT & 95.4 & 97.8 & 99.8 & 93.0 & 100.0 & 100.0 \\
PTv3 Regression & 74.1 & 56.2 & 51.4 & 26.7 & 31.8 & 22.3 \\
PlaceIt3D$^{\dagger}$ & \textbf{96.9} & 52.4 & 60.9 & 31.2 & 44.5 & 15.1 \\
\noalign{\vskip 3pt}
\hdashline
\noalign{\vskip 3pt}
Vanilla & 84.0 & 67.6 & 69.9 & 43.5 & 58.2 & 37.5 \\
GOPI-Single & 91.9 & 75.8 & 83.0 & 60.3 & 60.3 & \textbf{63.3} \\
GOPI-Dual & 92.9 & \textbf{80.8} & \textbf{85.7} & \textbf{67.3} & \textbf{69.5} & 62.3 \\
\specialrule{1pt}{5pt}{4pt}
\multicolumn{7}{c}{\textit{(b) Scene-disjoint split}} \\
\midrule[0.4pt]
PTv3 Regression & 73.0 & 53.8 & 53.2 & 23.9 & 33.1 & 22.9 \\
PlaceIt3D$^{\dagger}$ & \textbf{97.0} & 53.2 & 53.1 & 28.4 & 32.9 & 21.1 \\
\noalign{\vskip 3pt}
\hdashline
\noalign{\vskip 3pt}
Vanilla & 93.1 & 70.4 & 76.1 & 52.7 & 64.0 & 44.6 \\
GOPI-Single & 88.1 & 72.2 & 83.8 & 55.7 & 57.1 & 60.8 \\
GOPI-Dual & 93.4 & \textbf{79.1} & \textbf{84.2} & \textbf{63.5} & \textbf{67.2} & \textbf{61.8} \\
\bottomrule
\end{tabular}
\end{table*}

Table~\ref{tab:main_results} summarizes the quantitative results
on 3D pose estimation under the asset-disjoint and scene-disjoint splits.
We include \emph{GT} poses as a reference under the same feasibility metrics.
Because dataset layouts and reconstructed geometry may contain
imperfections, GT should not be interpreted as a strict upper bound.

\emph{PTv3 Regression} is a direct regression baseline that uses a PTv3 encoder followed by GeM pooling and MLP heads to predict furniture translation and yaw in a single forward pass.
This baseline represents the most straightforward formulation of the task, without iterative refinement or explicit generation-oriented geometric constraints.

\emph{PlaceIt3D$^{\dagger}$}\cite{placeit3d2025} is adapted to our task while retaining
its placement-point selection and discrete orientation prediction.
We provide scene and asset point clouds, specify the insertion
region using frustum points, and use a shared generic placement
instruction.
Candidate construction, including GT-assisted point supplementation
for the original and replacement settings, and training details
are provided in the supplementary material Adaptation of PlaceIt3D.

PlaceIt3D$^{\dagger}$ improves Overall@M over PTv3 Regression
from $26.7\%$ to $31.2\%$ under the asset-disjoint split and
from $23.9\%$ to $28.4\%$ under the scene-disjoint split.
It achieves the highest support validity among learned methods,
potentially aided by its support-related candidate construction
and GT-assisted supplementation.
However, its YawAcc@10$^\circ$ is lower than PTv3 Regression
under both splits ($15.1\%$ vs.\ $22.3\%$ and
$21.1\%$ vs.\ $22.9\%$, respectively).

\emph{Vanilla} denotes the basic implementation of GOPI, without
frustum points, the frustum consistency constraint, support alignment
loss, or room-axis alignment.
It outperforms PTv3 Regression across all metrics under both splits.
Under the asset-disjoint split, Overall@M increases from $26.7\%$
to $43.5\%$, and TransAcc@0.3m from $31.8\%$ to $58.2\%$.
Under the scene-disjoint split, the corresponding improvements
are from $23.9\%$ to $52.7\%$ and from $33.1\%$ to $64.0\%$.
These results indicate that the basic GOPI formulation improves
upon direct pose regression even when training and test scenes
are disjoint.

\emph{GOPI-Single} and \emph{GOPI-Dual} are two backbone variants
of the proposed framework.
GOPI-Single uses a single PTv3 encoder over the merged
furniture--environment point cloud, whereas GOPI-Dual uses
the proposed dual-stream object--environment design.
Both variants share the same iterative refinement and geometric
reasoning modules.
As shown in Table~\ref{tab:main_results}, both variants achieve
higher Overall@M than PTv3 Regression, PlaceIt3D$^{\dagger}$, and
Vanilla under both splits, supporting the effectiveness of
the proposed framework in jointly satisfying placement constraints.

Compared with GOPI-Single, GOPI-Dual improves Overall@M and
TransAcc@0.3m by $6.9$ and $9.2$ pp under the asset-disjoint split,
with only a $1.0$ pp decrease in YawAcc@10$^\circ$.
Under the scene-disjoint split, it improves all reported metrics,
including gains of $7.7$ pp in Overall@M and $10.1$ pp in TransAcc@0.3m.
Although PlaceIt3D$^{\dagger}$ achieves higher support validity,
potentially aided by its candidate-point supplementation,
GOPI-Dual requires no such supplementation and outperforms it
on all other reported metrics under both splits.
These results demonstrate the dual-stream design's advantage
in balancing joint feasibility and pose accuracy.

\begin{figure}[t]
    \centering
    \includegraphics[width=\linewidth]{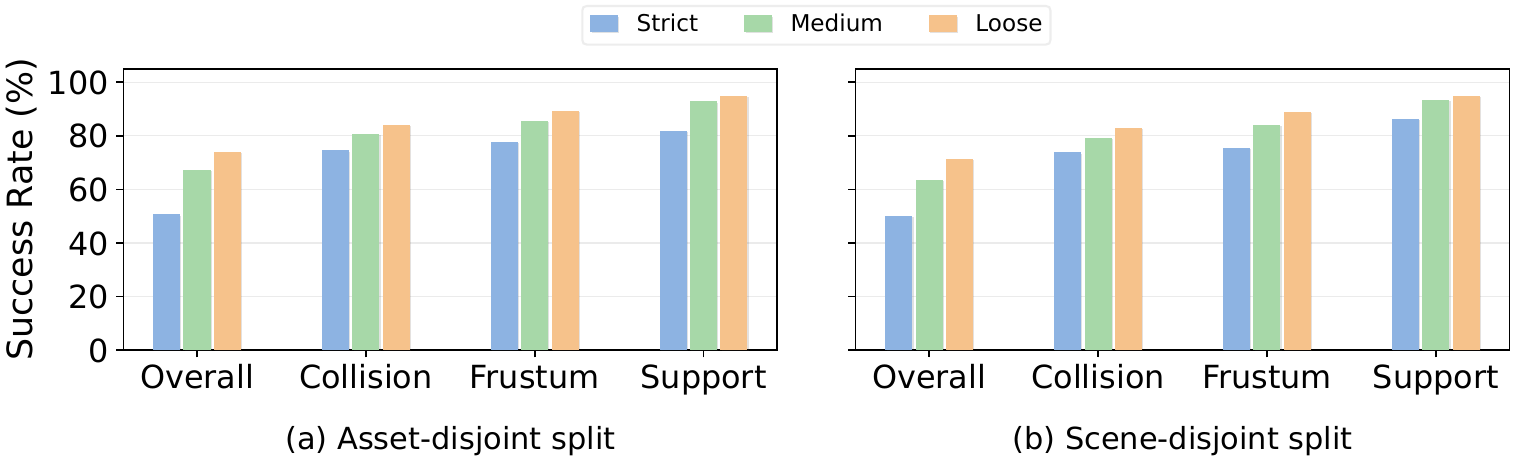}
    \caption{
    \textbf{Feasibility pass rates of GOPI-Dual under three threshold settings.}
    (a) Asset-disjoint split. (b) Scene-disjoint split.
    }
    \label{fig:pass_rates}
\end{figure}

Fig.~\ref{fig:pass_rates} compares GOPI-Dual under strict, medium,
and loose feasibility thresholds across both splits.
Support validity remains higher than collision and frustum validity,
indicating that the latter two constraints are the main bottlenecks.
Both splits exhibit similar trends, with success rates increasing
as the thresholds are relaxed.
Overall validity requires all three criteria to pass jointly
and therefore remains below the individual pass rates.

\subsection{Generalization to Novel Insertions}
\label{sec:replacement_placement}
We further evaluate GOPI beyond recovery of the original furniture
instance through same-category replacement, cross-category
replacement, and independent insertion masks
(Table~\ref{tab:replacement_results}).
Replacement assets are absent from the source scene and have
target-asset IDs disjoint from those used for training.
All methods use the same scene--asset pairs and insertion masks.

\paragraph{Same-Category Replacement.}
For each eligible test sample, we randomly select three replacement
assets from the same category, while preserving the support type
and maintaining similar bounding-box dimensions.
This yields 2,904 placements.
GOPI achieves an Overall@M of $64.2\%$, compared with $25.7\%$
for PTv3 Regression and $33.1\%$ for adapted PlaceIt3D.
Although PlaceIt3D achieves higher support validity, GOPI obtains
stronger collision and frustum validity, resulting in better joint
feasibility.
These results indicate that GOPI can accommodate changes in furniture
geometry within a category rather than relying on the original asset.

\paragraph{Cross-Category Replacement.}
We additionally replace the original furniture with assets from
different categories under the same support-type and size constraints,
resulting in 2,688 placements.
GOPI achieves an Overall@M of $64.2\%$, exceeding PTv3 Regression
and adapted PlaceIt3D by $35.8$ and $28.7$ percentage points,
respectively.
It also achieves the highest translation and yaw accuracy,
which here measure consistency with the original reference pose.
These results support the applicability of GOPI to new
scene--object combinations involving category changes.

\paragraph{Independent Insertion Masks.}
We evaluate 100 scene--asset pairs using manually drawn masks
that are not derived from the original furniture projection.
Since these masks do not define a unique reference placement,
we report feasibility metrics only.
GOPI achieves an Overall@M of $47.0\%$, compared with $39.0\%$
for both baselines, and obtains the highest collision and frustum
validity.
This demonstrates that its placement capability extends to
independently specified insertion regions.
Nevertheless, fewer than half of the placements satisfy all three
criteria jointly, indicating that this setting remains challenging.
Further details on asset selection and independent-mask annotation
are provided in the supplementary material Novel Insertion Settings.

\begin{table}[t]
\centering
\caption{
\textbf{Placement results with replacement assets and independent masks.}
PTv3 Reg denotes PTv3 Regression.
N/A indicates that no reference pose is defined.
}
\label{tab:replacement_results}
\footnotesize
\setlength{\tabcolsep}{2pt}
\renewcommand{\arraystretch}{1.12}
\resizebox{\columnwidth}{!}{%
\begin{tabular}{@{}lcccccc@{}}
\toprule
& \multicolumn{4}{c}{\textbf{Feasibility (\%) $\uparrow$}}
& \multicolumn{2}{c}{\textbf{Accuracy (\%) $\uparrow$}} \\
\cmidrule(lr){2-5}\cmidrule(lr){6-7}
\textbf{Method}
& \shortstack{Support\\@M}
& \shortstack{Collision\\@M}
& \shortstack{Frustum\\@M}
& \shortstack{Overall\\@M}
& \shortstack{TransAcc\\@0.3m}
& \shortstack{YawAcc\\@10$^\circ$} \\
\midrule
\multicolumn{7}{c}{\textit{(a) Same-category replacement}} \\
\midrule
PTv3 Reg & 75.0 & 55.6 & 52.3 & 25.7 & 31.3 & 23.5 \\
PlaceIt3D$^{\dagger}$ & \textbf{96.9} & 56.9 & 57.3 & 33.1 & 40.4 & 16.0 \\
GOPI & 90.8 & \textbf{82.7} & \textbf{83.3} & \textbf{64.2} & \textbf{65.6} & \textbf{62.5} \\
\specialrule{1pt}{5pt}{4pt}
\multicolumn{7}{c}{\textit{(b) Cross-category replacement}} \\
\midrule
PTv3 Reg & 80.4 & 57.6 & 50.9 & 28.4 & 35.8 & 19.0 \\
PlaceIt3D$^{\dagger}$ & \textbf{98.9} & 54.8 & 60.7 & 35.5 & 43.5 & 16.3 \\
GOPI & 95.5 & \textbf{80.4} & \textbf{78.9} & \textbf{64.2} & \textbf{65.4} & \textbf{62.6} \\
\specialrule{1pt}{5pt}{4pt}
\multicolumn{7}{c}{\textit{(c) Independent-mask replacement}} \\
\midrule
PTv3 Reg & 72.0 & 71.0 & 67.0 & 39.0 & N/A & N/A \\
PlaceIt3D$^{\dagger}$ & \textbf{94.0} & 67.0 & 58.0 & 39.0 & N/A & N/A \\
GOPI & 79.0 & \textbf{81.0} & \textbf{72.0} & \textbf{47.0} & N/A & N/A \\
\bottomrule
\end{tabular}%
}
\end{table}

\subsection{Analysis of Feasibility and Pose Accuracy}

\begin{figure}[t]
    \centering
    \includegraphics[width=1.0\linewidth]{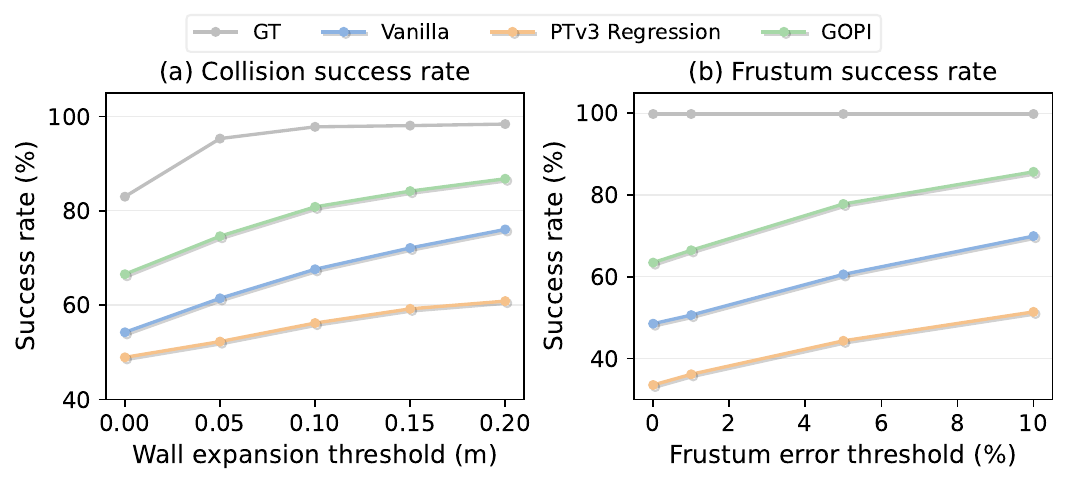}
    \caption{
\textbf{Comparison of feasibility success rates under varying thresholds.}
(a) Collision success rate as the room-boundary expansion distance varies.
(b) Frustum success rate as the frustum error threshold changes.
}
    \label{fig:feasibility_curves}
\end{figure}

Fig.~\ref{fig:feasibility_curves} compares collision and frustum success rates under varying thresholds.
GOPI consistently outperforms Vanilla and PTv3 Regression on both criteria, while Vanilla also remains stronger than PTv3 Regression.
This shows that the proposed framework improves geometric feasibility across a broad range of tolerance levels, rather than only under a single operating point.

\begin{figure}[t]
    \centering
    \includegraphics[width=1.0\linewidth]{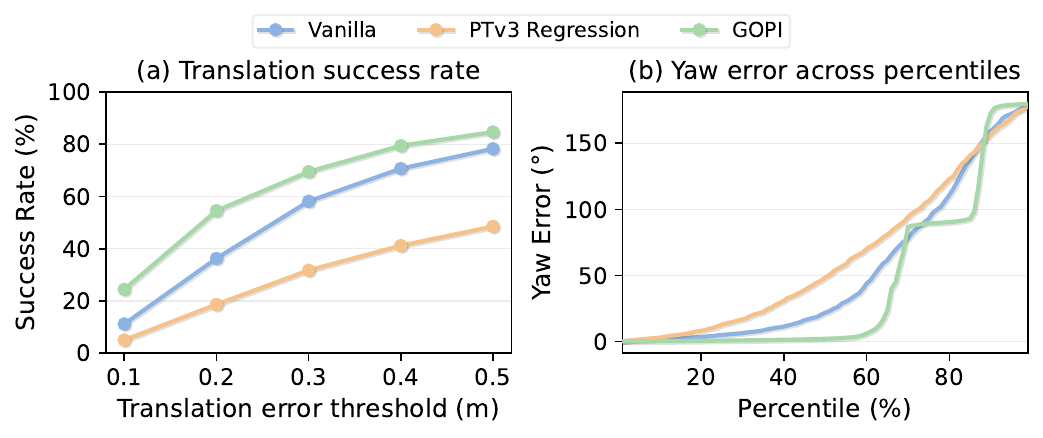}
    \caption{
\textbf{Comparison of pose accuracy against baseline methods.}
(a) Translation success rate as the translation error threshold varies.
(b) Yaw error distribution across percentiles.
}
    \label{fig:pose_compare_merged}
\end{figure}

Fig.~\ref{fig:pose_compare_merged} further analyzes pose accuracy.
As shown in Fig.~\ref{fig:pose_compare_merged}(a), GOPI consistently achieves higher translation success rates than Vanilla and PTv3 Regression across different translation error thresholds.
Fig.~\ref{fig:pose_compare_merged}(b) reports the yaw error distribution across percentiles.
Compared with the smoother curves of the baselines, GOPI exhibits a step-like yaw error profile.
This behavior is mainly caused by room-axis alignment, which pulls yaw predictions that are close to dominant room directions toward the estimated room axes.
As a result, small yaw deviations can be corrected more effectively, whereas predictions that are far from the correct axis may incur larger errors.

Overall, these results indicate that GOPI produces poses with both stronger geometric feasibility and better agreement with reference layouts than the baselines.

\subsection{Ablation Study}

Table~\ref{tab:ablation} analyzes the contribution of each component in GOPI.
Removing iterative refinement causes the largest degradation,
reducing Overall@M from $67.3\%$ to $32.4\%$ and
TransAcc@0.3m from $69.5\%$ to $39.8\%$.
This highlights the importance of iterative pose refinement
under partial observability.
The frustum-related components are also important:
removing the frustum consistency loss reduces Frus@M
from $85.7\%$ to $74.9\%$, while removing frustum points
substantially reduces support, collision, overall feasibility,
and translation accuracy.
Room-axis alignment in refinement (RAAR) mainly contributes
to orientation estimation, as removing it decreases
YawAcc@10$^\circ$ from $62.3\%$ to $47.5\%$.
Although certain ablated variants achieve slightly better scores
on individual metrics, the full GOPI model achieves the highest
Overall@M and TransAcc@0.3m, indicating the benefit of
combining these components.

\begin{table}[t]
\centering
\caption{\textbf{Ablation study of GOPI (\%).}
Each row removes one component from the full GOPI model.}
\label{tab:ablation}

\setlength{\tabcolsep}{6pt}
\renewcommand{\arraystretch}{1.1}
\begin{tabular}{lcccccc}
\toprule
\textbf{Method}
& \shortstack{Supp\\@M}
& \shortstack{Coll\\@M}
& \shortstack{Frus\\@M}
& \shortstack{Overall\\@M}
& \shortstack{Trans\\@0.3m}
& \shortstack{Yaw\\@10$^\circ$}
\\
\midrule
w/o Iteration & 73.7 & 57.4 & 70.7 & 32.4 & 39.8 & 30.6 \\
w/o FruLoss & \textbf{93.6} & 76.6 & 74.9 & 55.1 & 63.2 & \textbf{63.5} \\
w/o RAAR      & 92.7 & 77.1 & \textbf{85.7} & 62.9 & 69.4 & 47.5 \\
w/o FruPts    & 76.9 & 66.3 & 80.8 & 44.4 & 51.9 & 57.5 \\
GOPI          & 92.9 & \textbf{80.8} & \textbf{85.7} & \textbf{67.3} & \textbf{69.5} & 62.3 \\

\bottomrule
\end{tabular}
\end{table}

\subsection{Effect of Mask Boundary Conditions}
We further analyze the influence of mask boundary conditions on GOPI.
An open mask is defined as a mask whose bounding rectangle touches
at least one image boundary, while a closed mask denotes the
complementary case.
Open masks can provide weaker spatial constraints because the
insertion region may extend beyond the image boundary,
making the corresponding frustum constraint less informative.

\begin{figure}[t]
    \centering
    \includegraphics[width=1.0\linewidth]{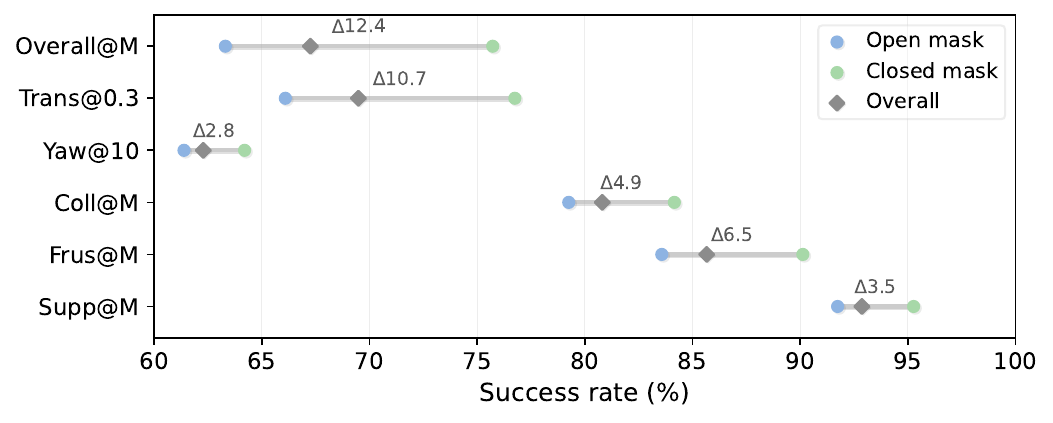}
    \caption{
\textbf{Open-mask vs. closed-mask performance comparison.}
The annotated $\Delta$ values show the closed--open performance gaps.
}
    \label{fig:gopi_open_closed_dumbbell}
\end{figure}

Fig.~\ref{fig:gopi_open_closed_dumbbell} compares GOPI on
1,041 open-mask and 486 closed-mask samples.
Closed masks yield higher scores across all reported metrics.
The largest gap occurs in Overall@M, which increases from
$63.3\%$ to $75.7\%$ ($12.4$ pp), followed by
TransAcc@0.3m with a $10.7$ pp gap.
Frustum validity also improves by $6.5$ pp.
These results are consistent with the greater difficulty of
localizing furniture and satisfying placement constraints
when insertion masks touch the image boundary.

\subsection{Evaluation of Image Generation}

\paragraph{Projection--Generation Consistency.}
We scale the furniture point cloud by factors of
$0.7$, $1.0$, $1.3$, and $1.6$, estimate its placement using GOPI,
and generate images conditioned on its projection.
We compare GOPI with projection conditioning against a 2D-only
baseline using the same generation backbone architecture without
the projection condition, and pretrained AnyDoor~\cite{anydoor2024}.
Specifically, the 2D-only baseline omits the four-channel
projection latent produced by the VAE encoder, reducing
the denoising network input from 13 to 9 channels.
All methods share the editing masks and are evaluated against
the same GOPI-derived target projections using Proj-Gen mIoU.
The 2D baselines do not receive these projections as generation inputs.

\begin{figure}[t]
    \centering
    \includegraphics[width=\linewidth]{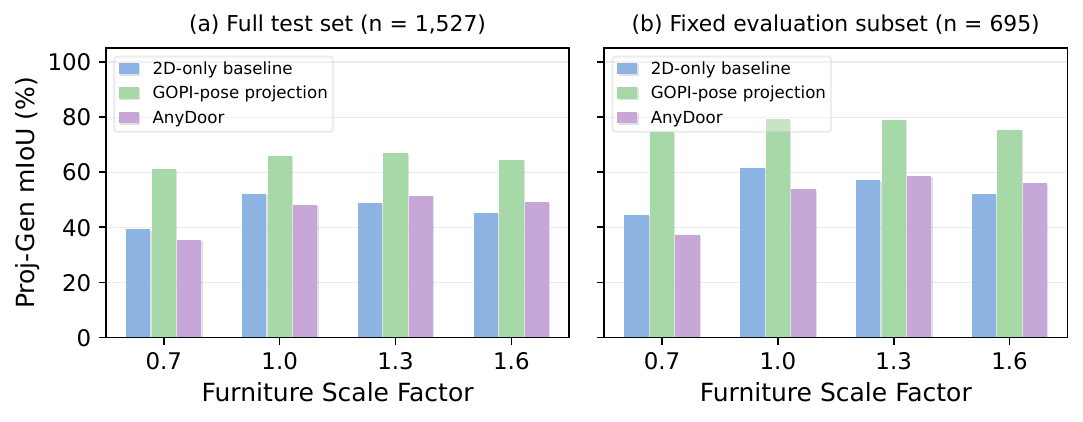}
    \caption{
    \textbf{Projection--generation consistency across furniture scales.}
    All methods share editing masks and target projections for evaluation.
    (a) Full test set ($n=1{,}527$).
    (b) Fixed evaluation subset ($n=695$).
    }
    \label{fig:mIoU}
\end{figure}

Fig.~\ref{fig:mIoU} reports results on all 1,527 test samples
and a fixed evaluation subset of 695 samples.
The subset contains samples for which the IoU between the
furniture mask segmented from the GT image and the corresponding
GT furniture projection exceeds $0.75$.
This selection reduces the influence of unreliable Grounded-SAM
segmentation and is fixed across all methods and scales.
The full-set mean retains all samples, assigning zero to
failed segmentations and empty reference projections.

GOPI achieves the highest Proj-Gen mIoU at every scale.
Averaged over four scales, its full-set score is $64.7\%$,
compared with $46.4\%$ for the 2D-only baseline and $46.0\%$
for AnyDoor; the corresponding subset scores are
$77.1\%$, $53.9\%$, and $51.4\%$.
These results support the benefit of explicit projection conditioning
beyond shared editing masks across the tested scales.

To examine whether projection conditioning remains effective under
both ground-truth and predicted poses, we compare 2D-only,
GT-pose-conditioned, and GOPI-pose-conditioned generation at scale
$\times1.0$, using the same editing masks and evaluation sets.
For GT-pose conditioning, we place the furniture at its
dataset-provided position and orientation, retain its original
3D scale, and render its projection using the original camera
parameters as the geometric condition for image generation.
The first two settings are evaluated against GT projections,
while the third is evaluated against GOPI-derived projections;
the 2D-only baseline receives no projection input.

Table~\ref{tab:pose_conditioning} shows that GT-pose conditioning
improves full-set Proj-Gen mIoU from $53.8\%$ to $67.6\%$,
while GOPI-pose conditioning achieves $66.0\%$ against
GOPI-derived projections.
Together with the matched comparisons in Fig.~\ref{fig:mIoU},
these results support effective projection conditioning under
both ground-truth and predicted poses.
Each conditioned output is evaluated against its input projection,
since a plausible placement need not coincide with the single
GT pose.
These scores measure adherence to the respective geometric
conditions rather than directly quantifying pose-estimation errors.

\begin{table}[t]
\centering
\caption{
\textbf{Projection--generation consistency at the original furniture scale.}
The reference projection used for evaluation is specified for each row.
Results are Proj-Gen mIoU (\%) on the full test set and the fixed subset.
}
\label{tab:pose_conditioning}
\setlength{\tabcolsep}{3pt}
\renewcommand{\arraystretch}{1.1}
\begin{tabular}{@{}llcc@{}}
\toprule
\textbf{Conditioning}
& \shortstack{\textbf{Reference}\\\textbf{projection}}
& \shortstack{\textbf{Full set}\\($n=1{,}527$)}
& \shortstack{\textbf{Subset}\\($n=695$)} \\
\midrule
2D-only              & GT   & 53.8 & 64.0 \\
GT-pose projection   & GT   & 67.6 & 81.8 \\
GOPI-pose projection      & GOPI & 66.0 & 79.5 \\
\bottomrule
\end{tabular}
\end{table}

\paragraph{Image Quality.}
We evaluate overall perceptual quality using full-image MUSIQ~\cite{Ke_2021_ICCV}
scores, as reported in Table~\ref{tab:image_quality}.
GOPI-pose-conditioned generation achieves an average MUSIQ score of $70.3$ across the four scales, closely matching the 2D-only baseline ($70.3$).
It also scores higher than AnyDoor at every tested scale.
Together with the alignment and human evaluation results,
these findings suggest that projection conditioning improves
control over generated furniture scale while maintaining
comparable overall image quality.

\begin{table}[t]
\centering
\caption{
\textbf{Image quality across furniture scales.}
Mean full-image MUSIQ scores on all 1,527 test samples
at each scale (KonIQ-10k weights; higher is better).
All generation methods share the editing masks.
Original dataset images are provided as a scale-independent reference.
}
\label{tab:image_quality}
\setlength{\tabcolsep}{4pt}
\renewcommand{\arraystretch}{1.1}
\begin{tabular}{@{}lcccc@{}}
\toprule
& \multicolumn{4}{c}{\textbf{MUSIQ $\uparrow$}} \\
\cmidrule(lr){2-5}
\textbf{Setting} & $\times0.7$ & $\times1.0$ & $\times1.3$ & $\times1.6$ \\
\midrule
Original dataset images & \multicolumn{4}{c}{72.2 (reference)} \\
\midrule
AnyDoor  & 68.1 & 68.0 & 67.5 & 66.7 \\
2D-only baseline & 70.4 & 70.4 & 70.3 & 70.3 \\
GT-pose projection & 70.3 & 70.4 & 70.2 & 69.9 \\
GOPI-pose projection & 70.4 & 70.5 & 70.3 & 70.1 \\
\bottomrule
\end{tabular}
\end{table}

\subsection{Human Evaluation}
\paragraph{Human Evaluation of 3D Placement.}
We evaluate perceived placement plausibility with 20 participants
on 100 cases randomly sampled from the full test set.
For each case, participants compare anonymized renderings of
the placements predicted by GOPI and adapted PlaceIt3D$^{\dagger}$ at the
original furniture scale.
Both methods use the same scene and target asset, and their
results are rendered with identical camera settings.
The target furniture is highlighted in blue, while the surrounding
scene is shown in gray.
Case order and method presentation order are randomized.
Participants judge the furniture's position, orientation, and
relationship with its surroundings, choosing which placement
is more reasonable or indicating that both are reasonable
or neither is reasonable.

\begin{figure}[t]
    \centering
    \includegraphics[width=\linewidth]{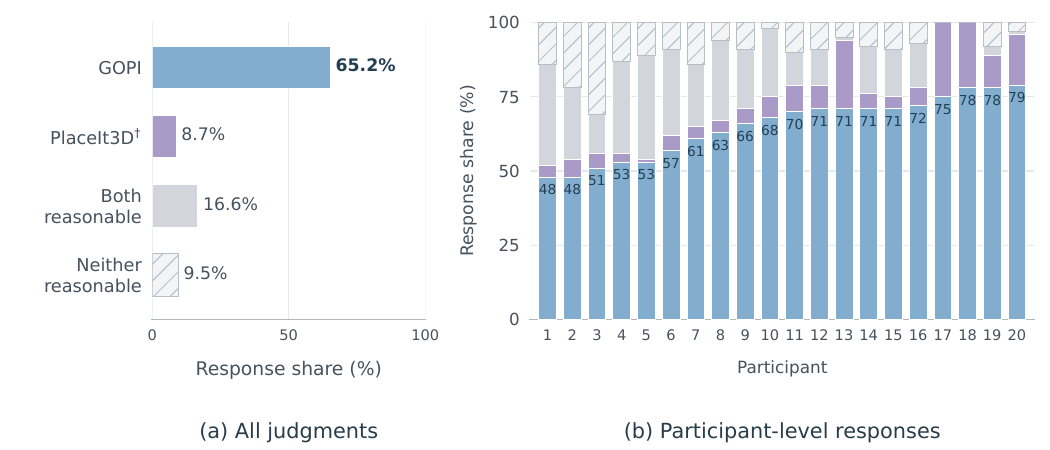}
    \caption{
    \textbf{Human evaluation of 3D placement plausibility.}
    (a) Overall distribution of the four response options.
    (b) Participant-level distributions, sorted by GOPI preference
    rate in ascending order.
    Numbers within the blue bars indicate GOPI preference percentages.
    Evaluation details are provided in the supplementary material.
    }
    \label{fig:pose_human}
\end{figure}

As shown in Fig.~\ref{fig:pose_human}, GOPI is preferred in
$65.2\%$ of the 2,000 judgments, compared with $8.7\%$ for
PlaceIt3D$^{\dagger}$.
Both placements are considered reasonable in $16.6\%$ of
judgments, while neither is considered reasonable in $9.5\%$.
All four response options are retained in the denominator.
GOPI receives more preferences than PlaceIt3D$^{\dagger}$ from every participant,
supporting its advantage in perceived placement plausibility.
The judgments that both placements are reasonable are consistent
with our discussion of observability and ambiguity:
different placements can be acceptable under the same partial
scene observation.
This human evaluation complements reference-pose accuracy
by assessing perceived placement plausibility beyond agreement
with a single reference pose.

\paragraph{Human Evaluation of Image Generation.}
We conduct a human evaluation on 100 cases randomly sampled from
the full test set, with 19 participants.
For each case, participants compare two anonymized image sequences
generated by GOPI and the 2D-only baseline using shared editing masks.
Each sequence contains four furniture scales
($0.7$, $1.0$, $1.3$, and $1.6$), with the same target projection
outlines overlaid for comparison.
Case order and method presentation order are randomized.
Participants select the sequence that better matches the target
outlines, with additional options for a tie or an uncertain judgment.

\begin{figure}[t]
    \centering
    \includegraphics[width=\linewidth]{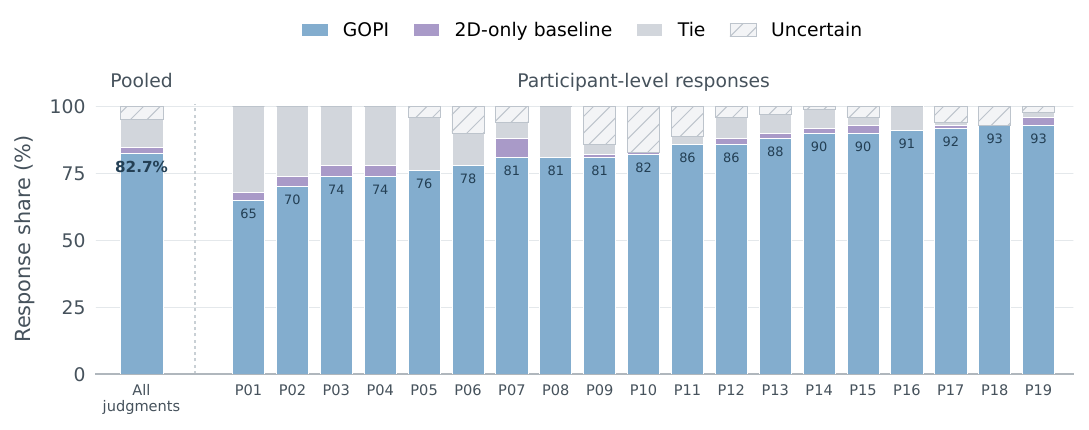}
    \caption{
    \textbf{Human evaluation of target-outline alignment.}
    Response distributions aggregated across participants (left)
    and shown per participant (right).
    Annotated values indicate GOPI preference percentages.
    Evaluation details are provided in the supplementary material.
    }
    \label{fig:human}
\end{figure}

As shown in Fig.~\ref{fig:human}, GOPI is preferred in $82.7\%$
of the 1,900 judgments, compared with $1.9\%$ for the baseline;
ties and uncertain judgments account for $10.7\%$ and $4.7\%$,
respectively.
GOPI preference rates range from $65\%$ to $93\%$ across participants,
with all responses retained in the denominator.
These results complement Proj-Gen mIoU by providing human evidence
that our method offers better control over the scale of generated
furniture than the 2D-only baseline within the tested scale range.

\subsection{Qualitative Results}
\begin{figure}[t]
    \centering
    \includegraphics[width=1.0\linewidth]{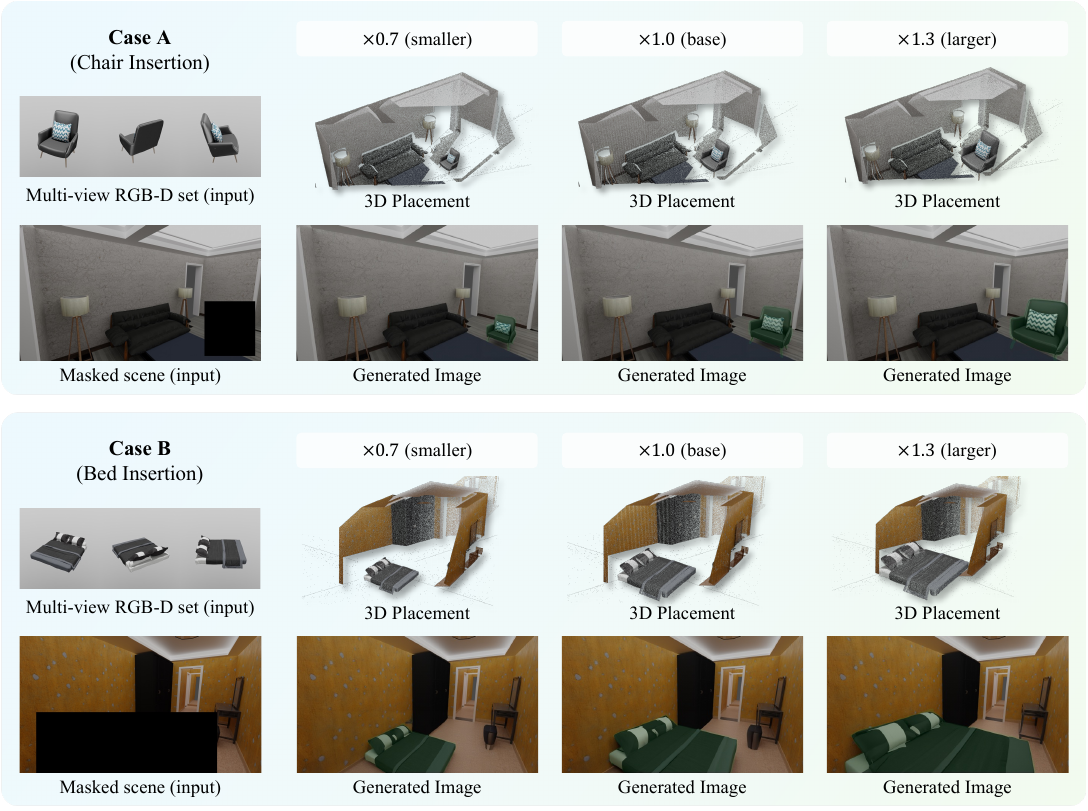}
    \caption{
    \textbf{Qualitative results under different furniture scale factors.}
    Green overlays denote the projected 2D masks of the inferred 3D furniture geometry.
    Zoom in for details.
    }
    \label{fig:vis}
\end{figure}

Fig.~\ref{fig:vis} presents qualitative examples of GOPI-guided furniture insertion under different furniture scale factors, including chair insertion in a living room and bed insertion in a bedroom.
For 3D placement, GOPI produces poses that are geometrically plausible and consistent with common furniture usage patterns.
For image synthesis, the green overlays denote the projected 2D masks of the inferred 3D furniture geometry.
Across different furniture scales, these projected masks remain closely aligned with the synthesized furniture regions, indicating that the generated images follow the geometry-aligned projection.

These examples further illustrate the advantage of the proposed two-stage design: by first inferring furniture placement in 3D and then conditioning image synthesis on the projected 3D geometry, the generated furniture preserves the geometry-defined scale and spatial extent in the image plane.

\section{Limitations}
Our current study has several limitations. First, all experiments are conducted on synthetic RGB-D indoor scenes, so real-world generalization remains to be established. Second, the current framework relies on depth observations for 3D placement inference, which may limit applicability in purely RGB settings. Third, we focus on single-object insertion and evaluate image generation primarily from a geometry-consistency perspective rather than as a comprehensive benchmark of visual realism.

\section{Conclusion}
In this paper, we studied furniture insertion in indoor scene images and argued that image-space masks alone do not uniquely determine physically plausible 3D placement under masked single-view conditioning.
To address this issue, we proposed a pose-first two-stage framework that first infers a geometrically plausible 3D furniture pose with GOPI and then uses its image-plane projection as a geometry-aligned condition for downstream image synthesis.
Experiments on synthetic RGB-D scenes show that the proposed framework improves geometric feasibility, better matches reference layouts under tolerance-based pose criteria, and maintains relatively stable projection--generation alignment across different tested furniture scales.
Overall, these findings support the central claim of this work: for geometry-consistent furniture insertion, it is beneficial to first solve 3D placement and then use the inferred geometry to constrain synthesis.

\bibliographystyle{IEEEtran}
\bibliography{GOPI}

\vfill

\end{document}